\documentclass[runningheads]{llncs}
\usepackage{xcolor}
\usepackage{graphicx}
\usepackage{hyperref}
\usepackage{upgreek}
\usepackage{amsmath}
\usepackage{amssymb}
\usepackage{biblatex}
\begin{document}
\title{Document Image Binarization in JPEG Compressed Domain using Dual Discriminator Generative Adversarial Networks}

\author{Bulla Rajesh\inst{1,2,*}\orcidID{0000-0002-5731-9755} \and
Manav Kamlesh Agrawal\inst{1,\Psi} \and
Milan Bhuva\inst{1,\Psi} \and
Kisalaya Kishore\inst{1,\Psi} \and
Mohammed Javed\inst{1,\Psi}\orcidID{0000-0002-3019-7401}
}
\authorrunning{Bulla. Rajesh et al.}
%
\institute{Department of IT, IIIT Allahabad, Prayagraj, U.P, 211015, Idia \and Department of CSE, Vignan University, Guntur, A.P, 522213, India \\
\email{\inst{*}br\_cse@vignan.ac.in}\\
\email{\inst{\Psi}\{rsi2018007,iit2018178,iit2018176,iit2018079,javed\}@iiita.ac.in}}

\maketitle              
\begin{abstract}
Image binarization techniques are being popularly used in enhancement of noisy and/or degraded images catering different Document Image Anlaysis (DIA) applications like word spotting, document retrieval, and OCR. Most of the existing techniques focus on feeding pixel images into the Convolution Neural Networks to accomplish document binarization, which may not produce effective results when working with compressed images that need to be processed without full decompression. Therefore in this research paper, the idea of document image binarization directly using JPEG compressed stream of document images is proposed by employing Dual Discriminator Generative  Adversarial Networks (DD-GANs).  Here the two discriminator networks - Global and Local work on different image ratios and use focal loss as generator loss. The proposed model has been thoroughly tested with different versions of DIBCO dataset having challenges like  holes, erased or smudged ink, dust, and misplaced fibres. The model proved to be highly robust, efficient both in terms of time and space complexities, and also resulted in state-of-the-art performance in JPEG compressed domain.

\end{abstract}

\section{Introduction}
\label{sec:intro}

\begin{figure}[!h]
        \center
		\includegraphics[width = 170pt, height = 100pt]{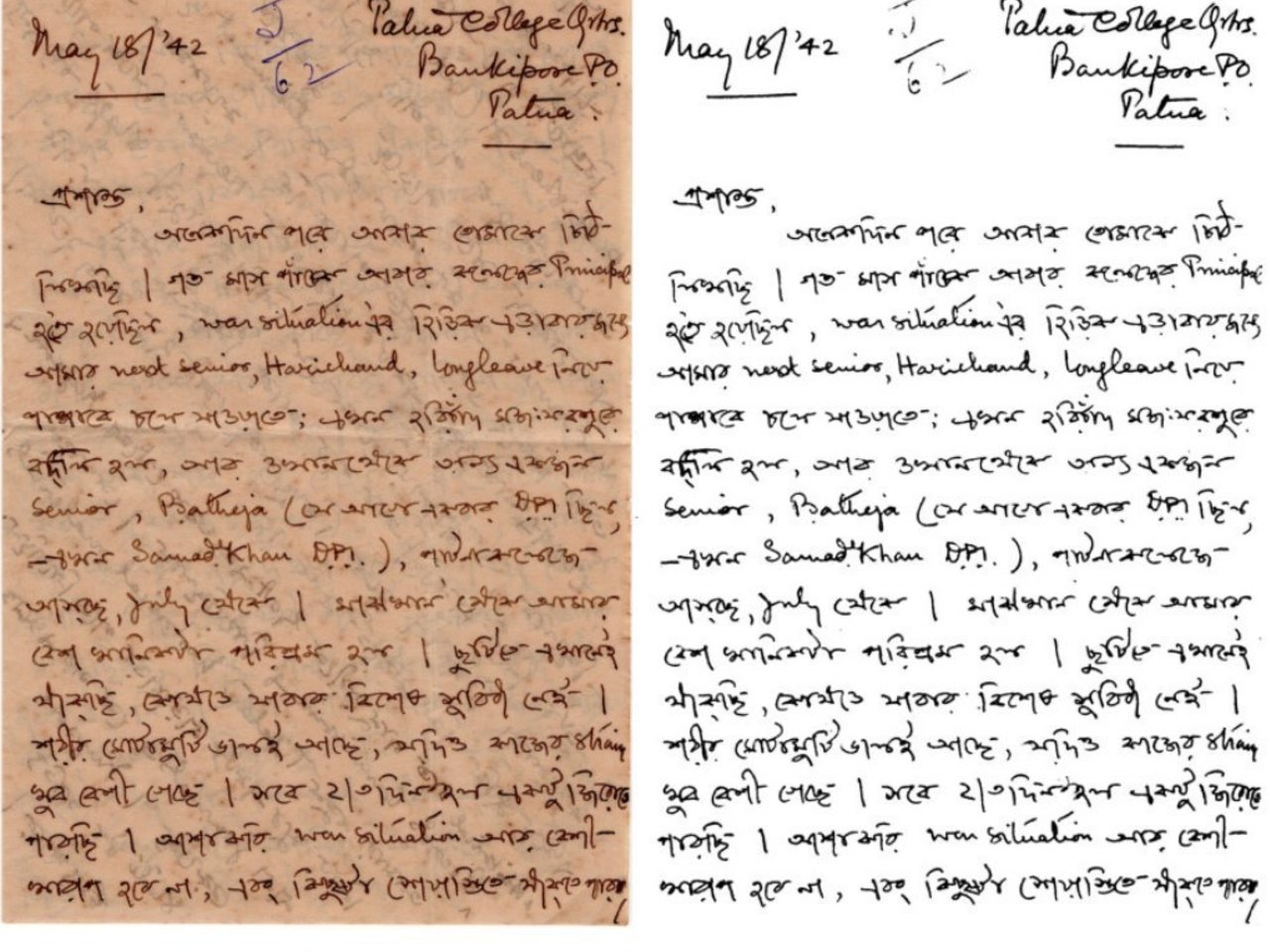}
			\caption{The problem of document binarization in case of a historical document}
			\label{samplemachnism}
\end{figure}

Document image binarization is a critical stage in any image analysis task, where eventually the image pixels are classified into text and background as shown in the Figure-\ref{samplemachnism}. This dominant stage can hamper recognition tasks in the later stages \cite{javed2019enhancement}. The need for this stage arises due to natural degradation of historical documents, such as aging effects, ink stains, bleed through, stamps and faded ink \cite{KHAMEKHEM_2022}. Moreover, digitized documents themselves might be compromised due to bad camera quality, disturbances, non-uniform illumination, watermarks, etc. These documents constitute plethora of information which could prove beneficial for us humans.
Therefore, in the literature there are plenty of research work focused on image binarization using both the handcrafted feature-based methods \cite{omar2019,javed2019enhancement} and deep learning-based methods \cite{inproceedings,dang2020,ayatollahi2013,rajonya2020}. These methods are pixel image driven, which may not be feasible when working with compressed document images that need to be processed without full decompression. This is because full decompression becomes an expensive task when huge volume of document images are to be processed. Therefore, in this research paper, the novel idea of image binarization using compressed document images is proposed that trains the deep learning model directly with the compressed stream of data.

In digitisation’s early stages, document binarization meant using single or hybrid thresholding techniques, such as Otsu method, Nick method, multi-level Otsu's method, and CLAHE algorithm \cite{omar2019}. However, with the advent of deep learning, CNNs' have been continually used to solve this rather taxing problem. Its superior performance over standard thresholding approaches can be due to its ability to capture the spatial dependence amongst the pixels \cite{DBLP:journals/corr/abs-1812-11690}.
The GANs’, deep learning technique, have been more successful than CNNs’ in the domain of image generation, manipulation and semantic segmentation \cite{suhSunghoReview}. GAN's such as Cycle GAN \cite{inbook}, ATANet coupled with UDBNet \cite{DBLP:journals/corr/abs-2007-07075}, conditional GAN \cite{inproceedings},and GAN with UNet \cite{dang2020}  can also combat the issue of limited data. Even conditional GAN's have been quite successful working on watermark removal from documents \cite{veeru2019}. Dual Discriminator GANs’, are a relatively new concept, have been used to remove degradation and they have proved to be robust and have performed on par with the current techniques. These deep learning networks have not only performed well on English based documents \cite{791885}, but also on Arabic and Hindi based documents \cite{6424434}. GAN's can also perform on Persian heritage documents \cite{ayatollahi2013}, and on hand-written Sudanese palm leaf documents \cite{8270066}. 

However, Dual Discriminator Generative Adversarial Network \cite{rajonya2020} are quite slow despite being shallow, owing to the presence of a heavy U-Net Architecture and 2 convolution neural networks. This issue can be solved to a certain extent by decreasing the size of the images, and we intend to use JPEG compression algorithm here. This algorithm intelligently employs Discrete Cosine transformation, quantisation and serialisation to remove redundancy of redundant features from an image. The human psychovisual system discards high-frequency information such as abrupt intensity changes and color hue in this type of compression. The accuracy of current techniques is frequently assessed through ICDAR conference \cite{inproceedings2}.

From the above context, the major contributions of this paper are given as follows- 

\begin{itemize}
    \item The idea of accomplishing document image binarization directly using JPEG compressed stream.
    \item The modified DD-GANs architecture with two discriminator networks - Global and Local working on different image ratios and using focal loss as generator loss, in order to accommodate JPEG compressed documents.
    \item Sate-of-the-art performance in JPEG compressed domain with reduced computation time and low memory requirements
   \end{itemize}

The rest of the paper is organized in 3 sections. Section 2 discuss the related literature, Section 3 briefs the proposed model and deep learning architecture. Section 4 reports the experimental results and presents analysis. Finally section 5 concludes the work with a brief summary and future work.

\section{Related Literature}
In this section, we present some of the prominent document binarization techniques that are reported in the literature. In \cite{omar2019}, Boudraa et al., have introduced a data pre-processing algorithm called CLAHE. This algorithm enhances visual contrast while preventing over-amplification of sound. However, applying the algorithm to all images is not a good strategy because it can modify the object boundaries and add distortion, affecting vital info in some circumstances. As a result, the contrast value is used. CLAHE is only done when the contrast value lies below a specific threshold.
In \cite{DBLP:journals/corr/abs-2010-08764}, the research goal is to improve the document images by tackling several types of degeneration. Some of these are removal of watermark(s), cleaning up of documents, and binarization. The purpose is to segment the image with watermark and the text in the foreground and background respectively. This is achieved using primarily detecting the watermark and then passing it through a designated model. 

The research work by \cite{rajonya2020}, proposes dual discriminator GAN. The architecture consists of U-Net as a generator and a  self designed local and global discriminator. Unlike normal GANs’, in dual discriminator GAN, local discriminator works on lower level features whereas global discriminator works on background features. Discriminators in normal GAN usually learn higher level or lower level features, but they cannot focus on both. Dual discriminators work better here as they can learn to recognise both.  Since the input documents contains more background noise, the focal loss function is used as the generator loss function to avoid the imbalance in the dataset, as suggested in the literature \cite{lin2017focal}, and binary cross entropy is used for the discriminator networks. Here they use a total loss function to reduce over-fitting in generator and discriminator. The local discriminator contributes more than the global discriminator to total loss as the local discriminator learns low level features which is to be reproduced by generator and thus is more important.

The work by \cite{7924246} revisits the formulation of JPEG algorithm. At the time of its creation, compression techniques such as predictive coding, block coding, cosine transformation, vector quantization, and combination of these were proposed. The Karhunen-Loeve Transform (KLT) introduced at the time, was the most optimum compression technique, however it was the most computationally intensive. DWT, other such technique, is also a more optimum compression technique, however, it too was not feasible with the hardware. Thus, Discrete cosine transform, which could be calculated very fast by using Fourier transforms, along with vector quantization and serialization with Zig-Zag encoding was decided as the major component of the algorithm. Also, the block size for compression was a major topic of discussion. Neither could it be so small that pixel-to-pixel correlation be missed, nor could it be large that block tries to take advantage of a correlation that might not exist. Thus the block size of 8$\times$8 was decided. Finally, the image was encoded using 2 algorithms simultaneously. The first pixel of each block is considered to be the AC component has Digital Pulse Code Modelling applied on it, whereas for the other 63 pixels, which are considered to be the DC component have run length encoding applied on them. 

There are some recent efforts to accomplish different DIA operations like text segmentation \cite{bulla2020,javed2018review}, \cite{rajesh2022fastss}, word recognition\cite{bulladcc} and classification \cite{bullacict}, and retrieval \cite{lu2003document} etc directly using the JPEG compressed stream. To the best of our knowledge, presently there is no image binarization technique available for JPEG compressed document images. Also JPEG is the most popular compression algorithm supported worldwide, and more than 90\% of images in the internet world are in JPEG format \cite{bulla2020}. Therefore, the image binarization method in this paper is focused only on addressing JPEG compressed document images. 

\section{Proposed Model}

\begin{figure*}[!h]
        \center
		\includegraphics[scale=0.55]{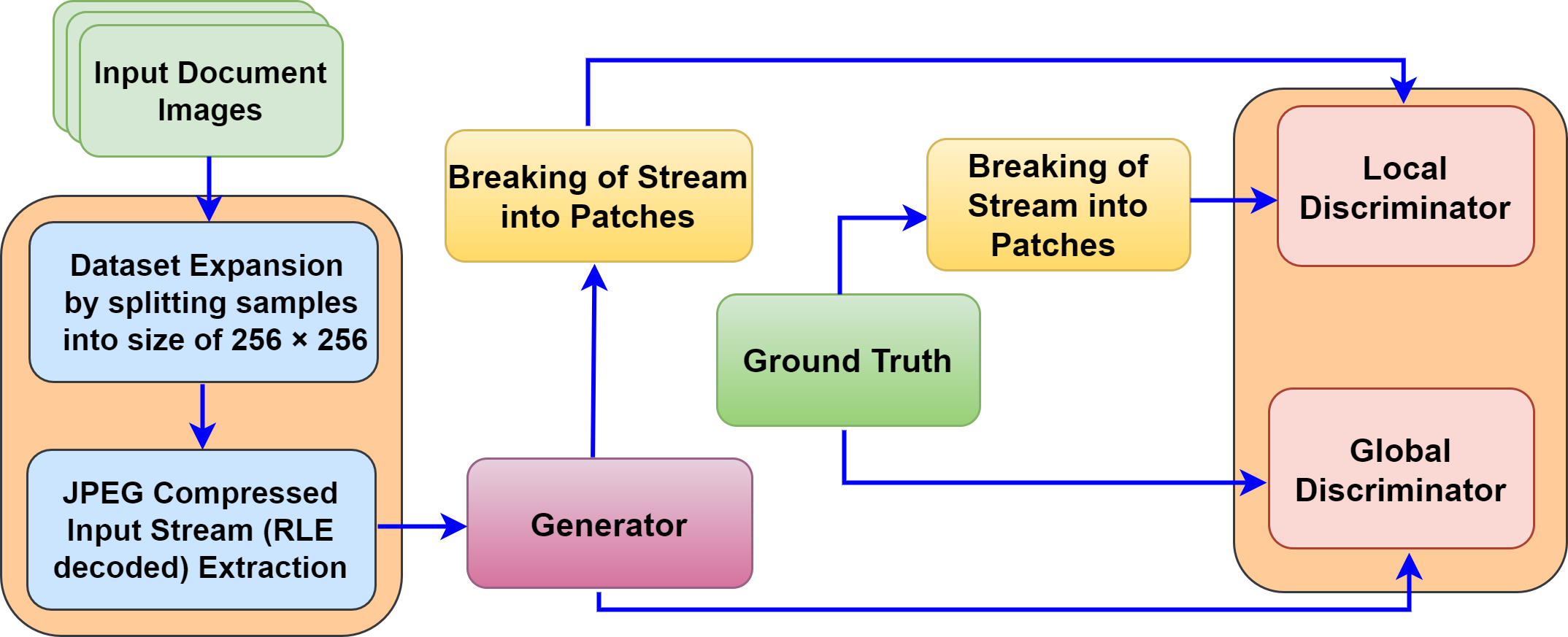}\par
			\caption{Proposed model for binarization of JPEG compressed document images. The deep learning architecture here uses average instead of max pooling layer, and it has 1 additional convolution layer in the local discriminator in comparison with the base model \cite{rajonya2020}}
			\label{fig1}
\end{figure*}

Our introduced model comprises of 2 major parts: pre-processing of pixel images, and DD-GAN as shown in Figure-\ref{fig1}. It is very important to note that, the pre-processing stage is required only for those documents that are not directly available in the JPEG compressed form. The input to the generator is always the JPEG compressed stream of document image to be binarized.  

\subsection{Image pre-processing}

We have used JPEG compression algorithm for pre-processing the images that are not made available in the JPEG compressed form. The JPEG algorithm takes advantage of the anatomical characteristics of the human eye. This method considers that humans are more sensitive to color illumination than chromatic value of an image, and that we are more sensitive to low frequency content in any picture than high frequency content.
JPEG algorithm consists of steps such as splitting of each image into blocks of size 8$\times$8, color-space transform, DCT, Quantisation, Serialisation, Vectoring, Encoding and Decoding. The step by step procedure of the algorithm is explained in \cite{SJPEG,bulla2020}.


\subsubsection{Splitting of image-}
The process of choosing the right block size, though might seem less significant, is one of the most significant part of the JPEG Algorithm. If we choose a small block size than we will not be able to find any relevant correlation between the pixels of the image, however a rather large size would lead to unnecessary advantage of a correlation that is not present. After careful consideration, the JPEG came up with the block size of 8$\times$8 for images of size 720$\times$575 or less. Since, each of the image that we have are of the size 256$\times$256, we use this block size.

\subsubsection{Color-Space Transform-}
The name suggests the process of this step in the algorithm. Here, a switch from RGB to YCbCr is made. The transformation is done for all the images in the RGB domain. We transform our image into this domain, as these colors are less sensitive to human eye and therefore can be removed. This color space is also more convenient as it separates the luminance and chrominance of the image.

\subsubsection{DCT-}
This mathematical transformation is a crucial step in JPEG compression algorithm.  Known as Discrete Cosine Transformation (DCT), it comprises of various mathematical algorithms such as Fast-Fourier Transform, which are used in turn to take any signal and transform it into another form. Since image is a type of signal, we can transform it into frequency or spectral information, so that it can be manipulated for compression using algorithms such as DPCM and Run Length Encoding(RLE). This transformation basically expresses each of those 8$\times$8 block pixels as sum of cosine waves. This helps us calculate the contribution of each cosine wave. As the high frequency contents of the image will have a less coefficient of cosine wave, we can remove those and only retain the low frequency ones. 

\subsubsection{Quantization-}
Quantization is the process where we remove the high frequency cosine waves while we retain the low frequency ones. For this, we have used the standard chrominance and luminance quantisation tables. These tables can be edited to change the JPEG compression ratio. 

\subsubsection{Serialisation-}
In serialisation we reduce the redundancy in the image, by zig-zag pattern scanning and serialise this data. Also, this groups the low frequency coefficients in the top of the vector.

\subsubsection{Vectoring-}
After applying DCT on the image, we are left 64 cosine waves, in which the first pixel is DC value whereas other 63 values are AC values. These DC values are large and can be varied but they will be similar to previous 8$\times$8 block, and thus is vectorised using Digital Pulse Code Modelling(DPCM). We also use Run Length Encoding to encode the AC components of the image. 

\subsubsection{Encoding-}
We use Huffman encoding technique to shrink the file size down further, and then the reverse the process to decode the image. However, full decoding is not necessary here in this research work, but partial decoding is needed to extract JPEG compressed DCT coefficients to be fed into the deep learning model. 

\subsection{Network Architecture}
The CNN used by us is Dual Discriminator based Generative Adversarial Network (DD-GAN) \cite{rajonya2020}. This model embraces a generator and a global and a local discriminator. A basic GAN model, vanilla GAN model, makes use of one of each generator and  discriminator, where the generator attempts to develop images that are plausible enough to fool the discriminator. These GAN's coupled with CNN's for feature extraction have made it plausible to rehabilitate ancient murals to a degree \cite{jianfang2020}. The main restriction in a single discriminator GAN model will be its limitation of having to choose between high level features or low level features, whichever gives a better plausibility to the model. And, since we ignore either one of them, it is plausible that the model is ignoring a rather significant feature of the image. Dual discriminator will allow each discriminator to extract both high level and low level features. The global discriminator is fed the entire image so that it extracts high level features like image background and texture, and the local discriminator, which is fed in patches of the entire image i.e. 32$\times$32 size of images extracted from 256$\times$256 image size, will extract low level features like the text strokes, edges, blobs .etc. 

\subsubsection{Generator -}
We use the U-Net architecture for the generator, proposed in the paper \cite{DBLP:journals/corr/RonnebergerFB15}. This generator architecture proposes down-sampling of the image, followed by up-sampling of same image. The down-sampling uses a typical convolution architecture of two 3$\times$3 convolution layers followed my a max pool having a stride of 2. The up-sampling of the image is done with a similar architecture as that of down-sampling architecture with the exception that the max-pooling layer is replaced with a up-convolution layer. Also, the generator uses the focal loss function. A major disadvantage of DIBCO dataset is the existence of pixels that belong to the background rather than text such as colored or white background, than the foreground pixels such as text strokes. This will create a largely imbalanced generator which will focus more on the background. This in turn will lead to a bad generator model. The focal loss function can treat this issue to some extent. This function petitions a modulating cross-entropy loss term in order to focus learning on hard miss classified instances and thus deals with class imbalance problem aptly.

\subsubsection{Discriminator -}
We use two discriminators: Global and Local. We have used binary cross entropy(BCE) loss as a local loss function for both discriminators. The global discriminator is made up of two convolution layers with batch normalization and the Leaky Relu activation function, followed by one average pooling layer and three fully connected layers. The global has a lesser number of layers than the local discriminator as it is supposed to deal with a background in the images, which consists of fewer features in ground truth images. The local discriminator is made up of 5 convolution layers coupled to batch normalization layers, and the activation function used is Leaky Relu. In addition, four fully connected layers are added to the discriminator. There is no average pooling layer as we consider patches of images of size 32$\times$32, and the use of a pooling layer might lead to loss of spatial information, which will be unfavorable as the discriminator is supposed to learn intricate features of the image. 

\subsection{Total GAN Loss}
The total GAN Loss tells how well the model is being trained. We use the focal loss from the discriminator and BCE loss from the discriminators in the following formula. The ~$\uplambda$ value is very high when compared to other values to signify that the generator is the main model being trained. 

\begin{itemize}
    \item The total loss function :
  \begin{gather*}\boxed{\emph{ L\textsubscript{total}} =\emph{ ~$\upmu$ (L\textsubscript{global} + ~$\upsigma$ L\textsubscript{local}) + ~$\uplambda$ \textsubscript{Lgen}}}
  \end{gather*}
  
    \item L\textsubscript{total} = Total loss, L\textsubscript{global} = global loss, L\textsubscript{local} = local loss averaged over all patches of images, Lgen = generator loss.
    \item The value of ~$\upmu$ given in the paper \cite{rajonya2020} is 0.5, 5, and 75. 
    \item The value of ~$\upsigma$ is higher than 1 to indicate that the global discriminator contributes lesser to the loss function than local .
\end{itemize}

\section{Experiment and Results}

\subsection{DIBCO Dataset}
\begin{figure}[!h]
        \center
		\includegraphics[ scale=0.06]{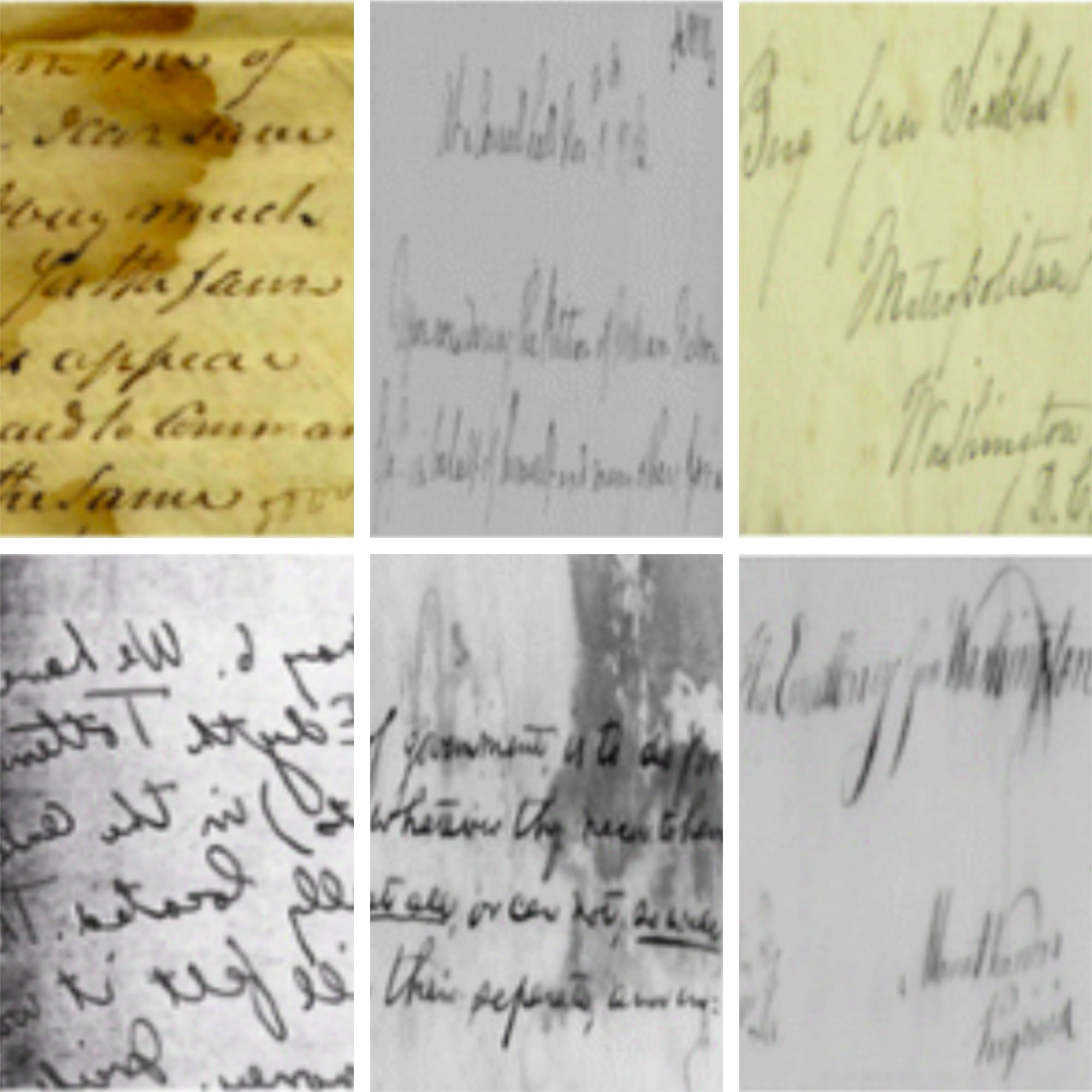}\par
			\caption{Some document images from DIBCO 09, 10, 11 Dataset \cite{hdibco2017}}
			\label{sampleimages}
			\vspace{-10pt}
\end{figure}

We have used the DIBCO-2014 \cite{hdibco2014}, 2016 \cite{hdibco2016}, 2017 \cite{hdibco2017} datasets to perform the experiment on the proposed model. DIBCO is a standardized dataset primarily used for document models which represent the challenges of binarization of historic handwritten manuscripts. Some of the sample document images are shown in the Figure-\ref{sampleimages}.

There are ten document images in the 2014 H-DIBCO, while H-DIBCO 2016 and 2017 consist of 20 document images each. DIBCO Dataset consists of document images as well as ground truth images which were built manually. This dataset poses the following challenges: Holes, erased and smudged ink, dust, and misplaced fibres, owing to 17th-century historical cloth documents. Furthermore, the digitization of these documents was done by institutions that own these documents and thus add non-uniformity in lighting, resolution etc., to the digitized documents. To escalate the size of the training and testing dataset, we pad each image with 128 black spaces on all sides and divide the image into blocks of 256$\times$256. This also allows us to train the model on less-resolution images, making the training faster. The train and test images of the entire dataset were passed through the JPEG algorithm. We got an average compression ratio of 20:1.

Firstly, we pre-process the DIBCO dataset images obtained from UCI machine learning repository \cite{Dua:2019} which are not directly available in the JPEG compressed format. The dataset consists of 20 documents and ground truth images. However, we cannot train a deep learning model with few images. Thus to expand the dataset, we divide each image into segments of 256$\times$256. This expansion creates sufficiently large training data of size 2352 images. After this, we pass the images through the JPEG compression algorithm to compress the images and feed them to the proposed DD-GAN Model. Primarily, We train a GAN by feeding the document images and ground truth of the document images. The document images are fed to the generator whilst the ground truth images are relegated to the discriminator. We feed the entire image (256$\times$256) to the global discriminator and patches of size 32$\times$32 into the local discriminator.
During testing also, we need to convert the testing image into compressed form, same as in training, before passing it through the generator. The generator will produce an image devoid of bleed-through, stain marks, uneven pen strokes, etc. Finally, we apply global thresholding of 127 to the generated image, giving us a binarized image. 

\subsection{Results}
\begin{figure}[!h]
        \center
		\includegraphics[width = 200pt, height = 200pt]{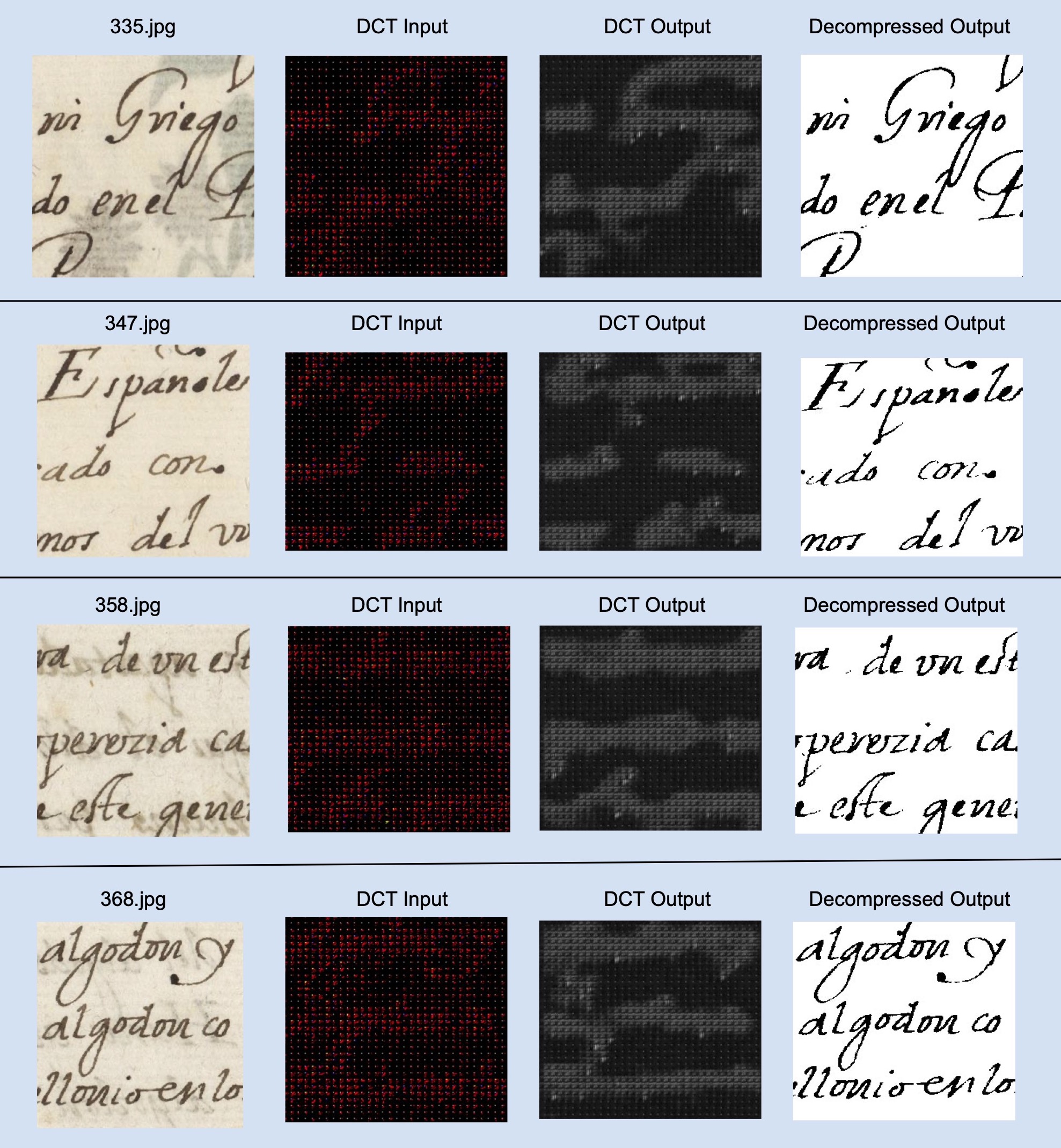}\par
			\caption{The output images generated by the proposed model for the sample input document images in the compressed domain. }
			\label{6}
\end{figure}

The experimental results of the proposed model tested on the standard dataset H-DIBCO have been tabulated in table \ref{2014d}, table \ref{2016d}, and table \ref{2017d}. 
We have employed the Peak Signal-to-Noise Ratio (PSNR) metric as the performance measurement as given in Eq (1).
\begin{equation}
\mathit{PSNR}={10\log_{10}{\frac{255^2}{MSE}}}
\end{equation}
The performance of the proposed model has been compared with the performance of the existing pixel domain model in the literature. In all the experiments, the proposed model has achieved better performance directly in the compressed domain, as shown in the tables. Similarly, we have calculated the performance in terms of pixel domain, where the PSNR value of the generated output of the proposed model is better when it is fully decompressed and compared with pixel domain output, as shown in the tables. Some of the output images in the compressed domain for the compressed input streams fed to the model are shown in Figure-\ref{6}. The middle two columns are the input stream and output stream in the compressed domain. In the figure, the uncompressed image of the compressed input stream of the model and the fully decompressed image of the compressed output stream computed by the proposed model are shown in the first and last columns, as shown in the Figure-\ref{6},  for human visual perception.

\begin{table}[!ht]
\centering
\caption{Test results on 2014 Dataset of H-DIBCO with the proposed model}
\begin{tabular}{c|c} 
 \hline
 \textbf{\small{Procedure}} & \textbf{\small{PSNR(\%)}} \\ 
 \hline
 Base model \cite{rajonya2020} & 22.60 \\
 \hline
 Proposed model with compressed JPEG images& 23.57 \\
 \hline
 Proposed model with fully Decompressed Images& 24.73 \\
 \hline
\end{tabular}
\label{2014d}
\end{table}

\begin{table}[!t]
\centering
\caption{Test results on 2016 Dataset of H-DIBCO with the proposed model}
\begin{tabular}{c|c} 
 \hline
 \textbf{\small{Procedure}} & \textbf{\small{PSNR(\%)}}  \\ 
 \hline
 Base model \cite{rajonya2020} & 18.83 \\
 \hline
Proposed model with compressed JPEG images & 19.64 \\
 \hline
Proposed model with fully decompressed images & 20.51 \\
 \hline
\end{tabular}
\label{2016d}
\end{table}

\begin{table}[!t]
\centering
\caption{Test results on 2017 Dataset of H-DIBCO with the proposed model}
\hspace{-2pt}\begin{tabular}{c|c} 
 \hline
 \textbf{\small{Procedure}} & \textbf{\small{PSNR(\%)}}  \\ 
 \hline
 Base model \cite{rajonya2020} & 18.34 \\
 \hline
 Proposed model with compressed JPEG images& 18.76 \\
 \hline
 Proposed model with fully decompressed images& 19.79 \\
 \hline
\end{tabular}
\label{2017d}
\end{table}

\begin{figure}[!ht]
	\vspace{-10pt}
        \center
		\includegraphics[scale=0.40]{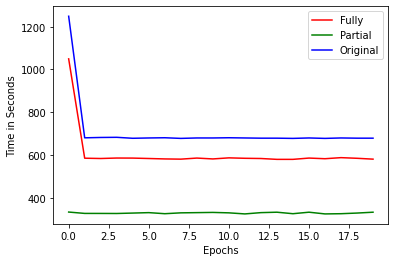}\par
			\caption{Time Analysis of proposed model based on raw images (original), JPEG compressed and fully decompressed images }
			\label{timegraph}
			\vspace{-10pt}
\end{figure}

Further, the advantage of the proposed model applying to direct compressed stream is in two folds. One is computational gain, and the second one is storage efficiency. We have conducted an experiment on the proposed model to verify these two advantages. The details are tabulated in table \ref{table4}. The model takes an average time of 708 seconds to run one epoch for pixel images and 333 seconds for compressed images, as shown in the table. Similarly, the storage cost of the compressed input stream is 48 Kb, which is very low compared to the uncompressed stream with a size of 3072 Kb, as shown in the table. Similarly, we have extended this experiment to the entire dataset, where the computational analysis for n number of images vs. the computational time it requires in both compressed and pixel domains is graphically shown in Figure-\ref{timegraph}. In both cases, the compressed input has shown reduced costs and improved performance in both computational and storage costs.

\begin{table}[!ht]
\centering
\caption{The performance details of the proposed model in terms of computational time and storage costs.}
\begin{tabular}{c|c|c} 
 \hline
 \textbf{\small{Procedure}} & \textbf{\small{Time/Epoch}} & \textbf{\small{Space/Batch}}  \\ 
 \hline
 Base Model \cite{rajonya2020} & 708sec & 3072kb\\
 \hline
 Proposed Model & 333sec & 48kb \\
 \hline
\end{tabular}
\label{table4}
\end{table}

\begin{figure}[!ht]
        \center
		\includegraphics[width = 250pt, height = 210pt]{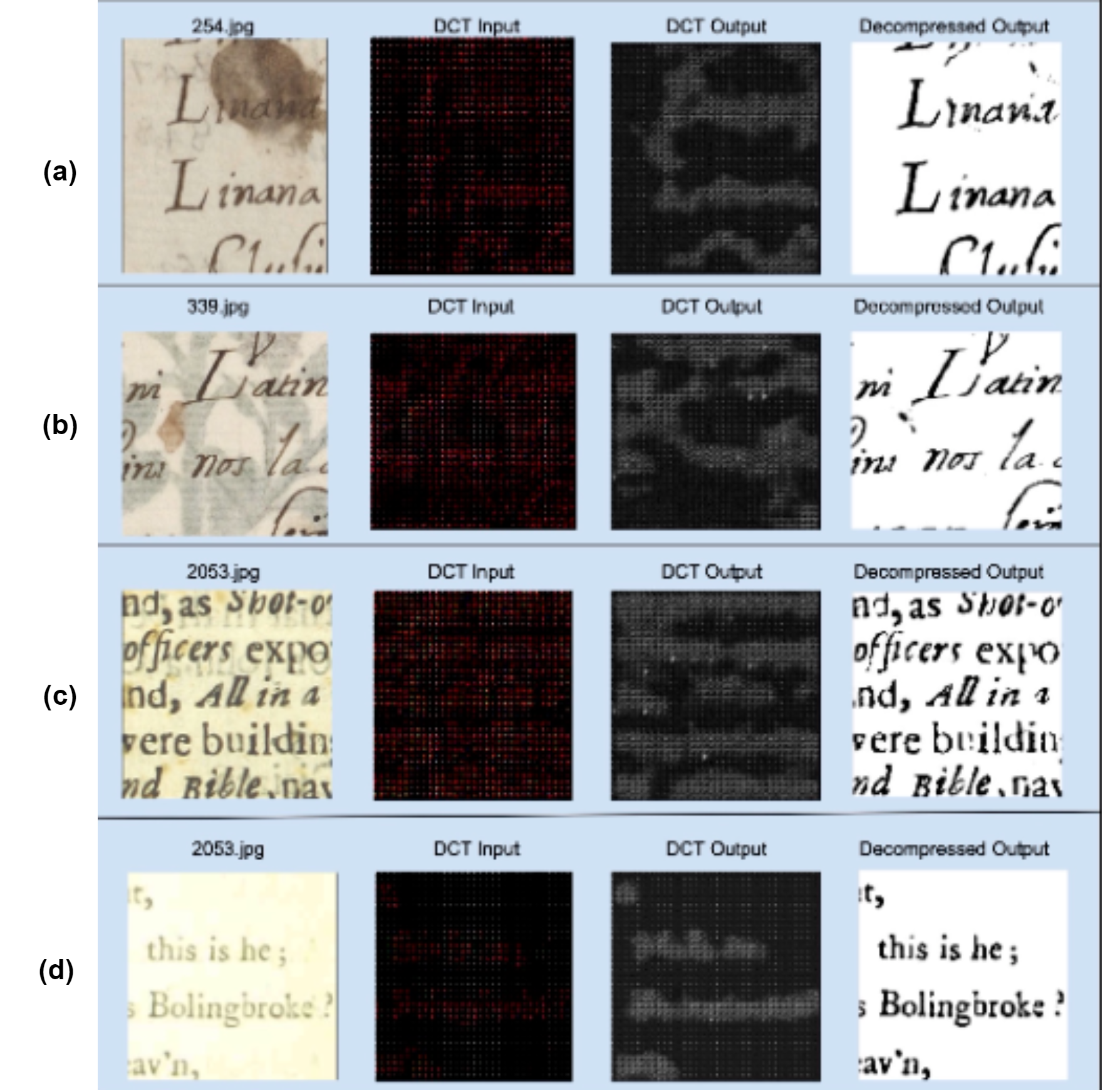}
			\caption{The experimental results of the proposed model tested on different challenging cases like (a) Smudged Ink, (b) Hole, (c) Dust, and (d) Uneven Lighting present in the dataset. }
			\label{fig8}
\end{figure}

\begin{figure}[!ht]
        \center
		\includegraphics[scale=0.53]{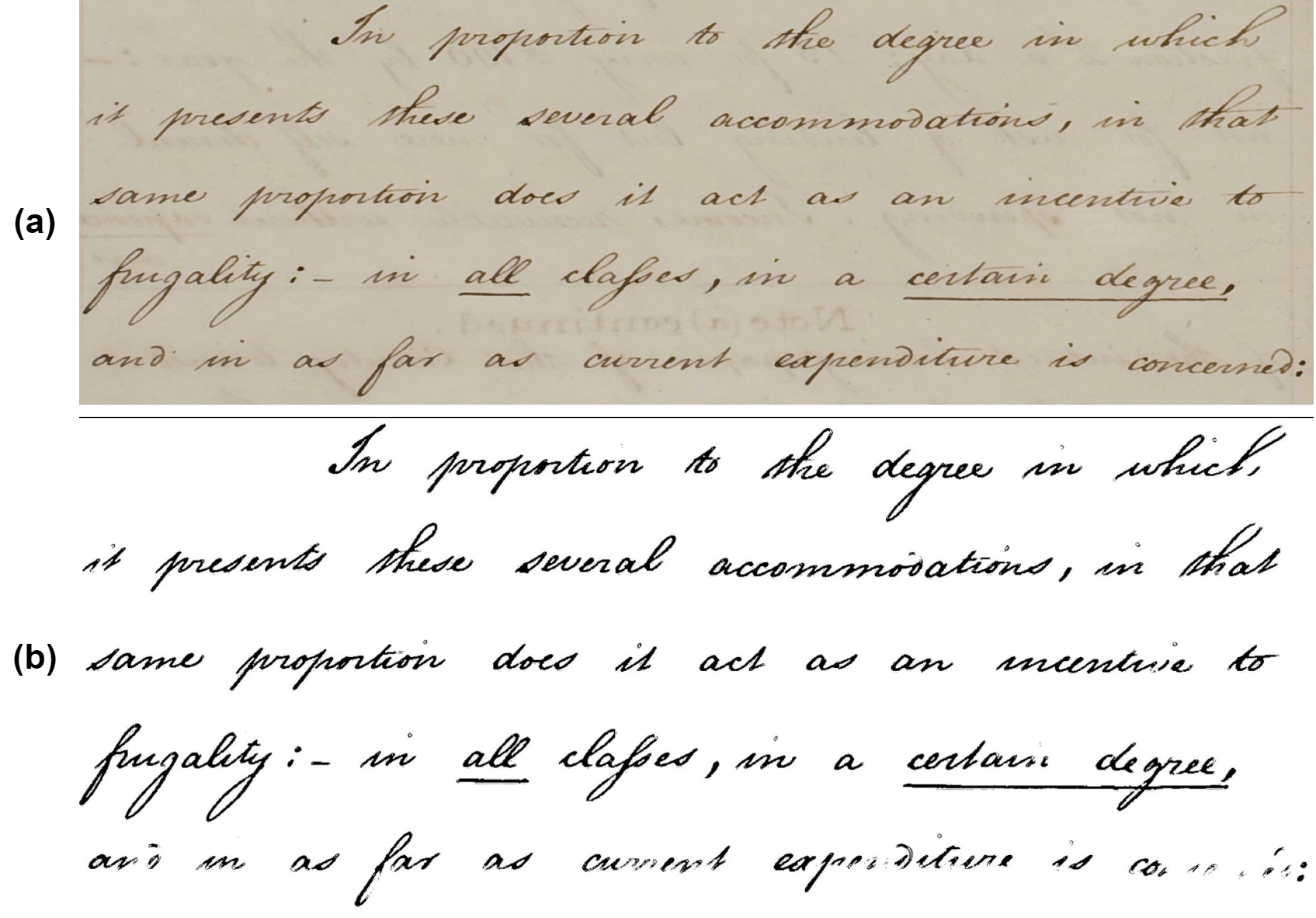}\par
			\caption{The experimental of the model tested on the entire document image showed in the decompressed domain (Row-wise), (a) Sample Input image with noise, (b) Predicted binarized image. }
			\label{fig9}
\end{figure}

The proposed model has been tested on different challenging input cases such as erased or smudged ink, uneven lighting, holes, and dust, present in the H-DIBCO Dataset. The output images of such cases are shown in Figure-\ref{fig8}. In the figure, the first row shows the output for the smudged ink. The second row shows the holes, and the third row shows the dust, and finally, the final row shows the results of uneven lighting. In all the cases, it can be observed that the proposed model has achieved significant performance directly in the compressed domain.
During training, since the proposed model was only trained on small patches of the input document images, we tested the model by verifying it on entire document images. Therefore, the proposed model may be ensured that it is not just limited to the small patches of the document image but also applicable to processing the entire document images. The experimental results on the entire document images are shown in the Figure-\ref{fig9}.
In the context of all the experiments discussed above, the overall observation and conclusion are that the proposed model in the compressed domain has proved to be an effective solution for binarizing document images directly in the compressed domain.  

\section{Conclusion}
The present research paper proposed a  model for performing the document image binarization using a dual discriminate generative adversarial network. The contribution of this research work is that the direct compressed stream of document images is fed to the proposed model to perform the binarization task directly in the compressed representation without applying decompression. The model has been tested on the benchmark dataset DIBCO, and the experimental results of the proposed model have shown the promising and stat-of-the
art performance directly in the compressed domain.
\printbibliography

\end{document}